\documentclass[10pt,twocolumn]{article}
\usepackage[margin=0.77in,columnsep=0.25in]{geometry}
\usepackage[T1]{fontenc}
\usepackage{mathptmx,microtype}
\usepackage{graphicx,booktabs,multirow,amsmath,amssymb,xcolor}
\usepackage{array,tabularx,siunitx,enumitem}
\usepackage[numbers,sort&compress]{natbib}
\usepackage[colorlinks=true,linkcolor=blue!45!black,citecolor=blue!45!black,urlcolor=blue!45!black]{hyperref}
\usepackage[capitalise]{cleveref}
\usepackage{placeins,multicol}
\usepackage{tikz}
\usetikzlibrary{arrows.meta,positioning,shapes.geometric,fit}
\setlist[itemize]{leftmargin=*,topsep=2pt,itemsep=1pt,parsep=0pt}
\setlist[enumerate]{leftmargin=*,topsep=2pt,itemsep=1pt,parsep=0pt}
\newcommand{\fig}[2]{\includegraphics[width=#2]{figures/#1.pdf}}

\title{Serving Agentic Robot Fleets:\\Workload Characterization, KV-Cache Thrashing,\\and Admission Control}
\author{Sida He\\[2pt]{\small Huawei Inc., China}}
\date{November 2026}
\hypersetup{pdftitle={Serving Agentic Robot Fleets: Workload Characterization, KV-Cache Thrashing, and Admission Control}, pdfauthor={Sida He}}
\begin{document}
\maketitle
\begin{abstract}
Agentic robots combine a vision--language brain, a tool-using harness, and a vision--language--action (VLA) policy. Serving a fleet of these agents introduces two coupled resource constraints: the brain repeatedly revisits long, growing key--value (KV) cache states, while the VLA executes action chunks under tight control deadlines. We characterize 80 LIBERO-PRO episodes generated by GPT-6-Astra, RPent, and $\pi_{0.5}$ (1,862 brain turns and 1,582 VLA inference chunks), profile both layers on H800 graphics processing units (GPUs), and replay the resulting agent programs on a vLLM serving engine. Brain contexts reach 215k tokens; the median Astra request takes 7.78\,s, while the intervening environment-step gap is only 1.0\,s. Under a 0.25\,s action-chunk deadline, one VLA GPU meets the 1\% starvation target for one robot; under a 1.0\,s deadline, its measured capacity reaches at least 32. In sustained brain replay, increasing concurrency from 20 to 32 robots reduces throughput from about 2.4 to 0.6 requests/s as repetitive prefill displaces generation. A static active-episode cap restores 2.24--2.25 requests/s and raises completed trace-program throughput from 1.30 to 5.00--5.55 episodes/min. Our measurements expose the workload-dependent shift between action deadlines and brain KV residency, and establish practical principles for capacity planning and admission in sustained agentic robot fleets.
\end{abstract}

\section{Introduction}
\label{sec:intro}

The computing architecture of a robot is changing. A capable vision--language model (VLM) can serve as a \emph{brain}: it observes the scene, reasons about task progress, and selects tools. A vision--language--action (VLA) policy such as $\pi_{0.5}$~\citep{pi05} transforms an instruction into a short sequence of motor actions. A coding-agent-style harness connects the two: the brain chooses a tool, receives its result, extends its context, and chooses the next action. Serving these brains from a shared graphics processing unit (GPU) cluster introduces long-lived conversational key--value (KV) cache state into robotic fleet scheduling.

Robotics serving research has established deadline-aware action-model scheduling, multi-model execution, and resource sharing~\citep{rosa,robion,beyondllm}. Agent serving research has established the importance of KV retention, tool-return timing, and thrashing under concurrent workloads~\citep{continuum,concur,cachewise,efficientagent,tao}. Their combination raises a new fleet-capacity question: \emph{how do the deadline of an action chunk and the working set of a frequently invoked, tool-driven brain jointly bound the number of robots a shared cluster can serve?}

\begin{figure}[t]
\centering
\resizebox{\columnwidth}{!}{%
\begin{tikzpicture}[font=\small,>=Latex,
  box/.style={draw,rounded corners=2pt,align=center,minimum width=31mm,minimum height=11mm,inner sep=3pt},
  service/.style={box,fill=blue!5},
  flow/.style={->,line width=.75pt},
  label/.style={font=\scriptsize,fill=white,inner sep=1pt}]
 \node[box] (robot) at (0,0) {Robot program\\sensing + tools};
 \node[service] (brain) at (5.4,1.35) {Brain serving\\growing KV state};
 \node[service] (vla) at (5.4,-1.35) {VLA serving\\action chunks};
 % Exact parallel lanes: each brain lane has dy/dx=+1.17/2.30,
 % and each VLA lane has dy/dx=-1.17/2.30. Ports stay on box edges.
 \draw[flow] ([yshift=4.8mm]robot.east) --
   node[label,pos=.46,above=1pt,sloped]{Context request}
   ([yshift=3.0mm]brain.west);
 \draw[flow] ([yshift=-0.8mm]brain.west) --
   node[label,pos=.54,below=1pt,sloped]{Tool decision}
   ([yshift=1.0mm]robot.east);
 \draw[flow] ([yshift=-1.0mm]robot.east) --
   node[label,pos=.46,above=1pt,sloped]{VLA tool call}
   ([yshift=0.8mm]vla.west);
 \draw[flow] ([yshift=-3.0mm]vla.west) --
   node[label,pos=.54,below=1pt,sloped]{Action result}
   ([yshift=-4.8mm]robot.east);
 \node[font=\scriptsize,align=center,text=black!65] at (5.4,-2.35) {Separately profiled GPU pools};
\end{tikzpicture}}
\caption{The embodied-agent program issues growing-context requests to the brain and action-chunk requests to the vision--language--action (VLA) service. Both layers are profiled on graphics processing units (GPUs); dependent brain requests are replayed on vLLM.}
\label{fig:system}
\end{figure}
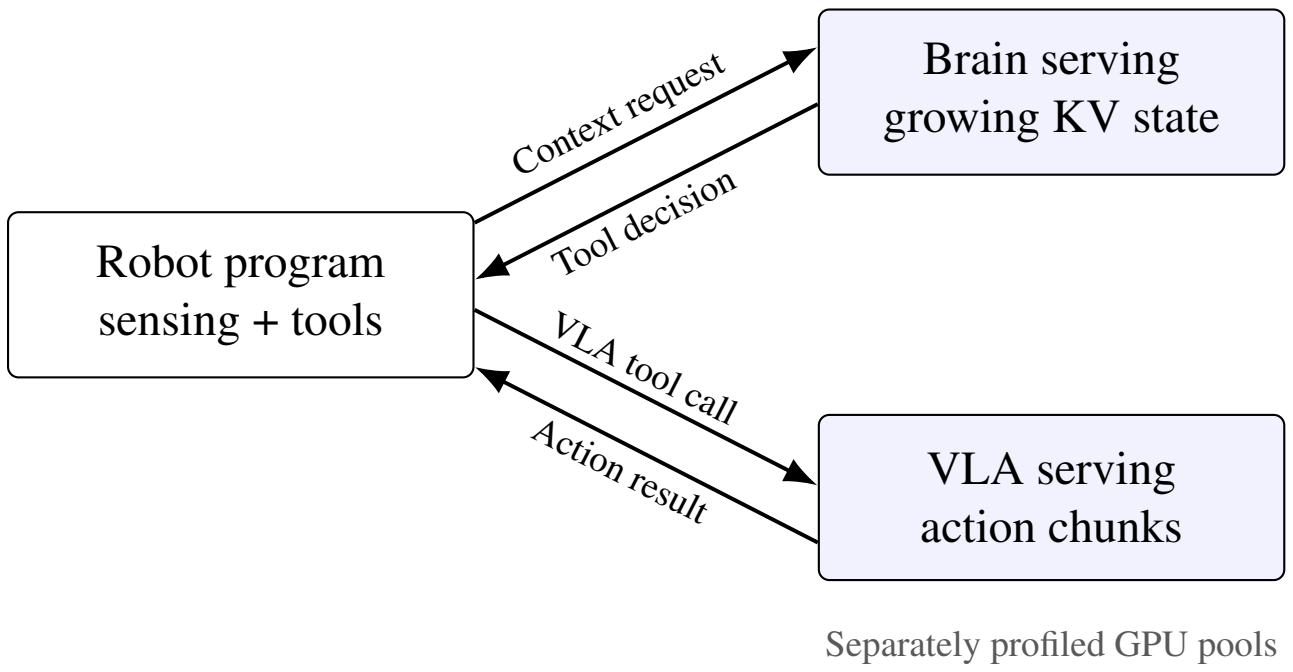

We answer with an empirical study of the robot programs and of the serving resources they consume. We collect 80 LIBERO-PRO episodes from GPT-6-Astra running the RPent harness with $\pi_{0.5}$. The traces contain 1,862 brain turns, 216 VLA-tool invocations, and 1,582 VLA inference chunks. We profile VLA latency and deadline-limited capacity, measure the compute cost of a Qwen3.6-27B brain proxy on H800 GPUs, and replay the measured growing-context request sequences on a real vLLM large language model (LLM) serving engine. \Cref{fig:system} summarizes the two resource constraints and the measurement design.

\noindent\textbf{Finding 1: the brain is a frequent, long-lived participant in control.} The median brain context is 78k tokens and the maximum reaches 215k. Brain turns occur 8.6$\times$ as often as VLA-tool entry points. The measured Astra request latency is 7.78\,s at the median, while the median intervening environment-step gap is 1.0\,s. A high-level agent can therefore spend substantially more time reasoning than its robot spends advancing between calls. This creates frequent reuse opportunities for growing KV states.

\noindent\textbf{Finding 2: robot capacity moves between deadline and KV constraints.} At a five-step, 20\,Hz action deadline of 0.25\,s, the measured one-GPU VLA capacity under a 1\% starvation service-level objective (SLO) is one robot. At a 1.0\,s deadline, the tested capacity reaches at least 32. Meanwhile, retaining a 130k-token brain prefix reduces the modeled 150-token call time from about 6.8\,s to 1.7\,s. The deployment envelope depends on both action deadlines and brain KV residency.

\noindent\textbf{Finding 3: sustained fleet demand exposes KV thrashing and the value of admission.} With a continuously replenished fleet, brain serving throughput falls from approximately 2.40 requests/s at 20 virtual robots to 0.60 requests/s at 32, while computed prompt tokens rise to about 1.5 million/min. Applying established working-set admission at episode boundaries restores 2.24--2.25 requests/s and raises complete trace-program throughput from 1.30 to 5.00--5.55 episodes/min. Finite-job comparison and transient overload tests further characterize how load duration and cache-recovery behavior shape this operating regime.

The contributions are \textbf{(1)} a turn-resolved characterization of embodied agents' context growth, tool timing, and brain latency; \textbf{(2)} an experimentally grounded two-layer fleet-capacity envelope; and \textbf{(3)} a sustained-fleet KV-thrashing analysis with admission benefits measured at both server-request and complete-program levels. Together these findings establish the resource model and telemetry required to serve long-running agentic robot fleets.

\section{Background and Related Work}
\label{sec:background}

\paragraph{Embodied brain--VLA systems.}
RPent exposes perception, geometric motion primitives, and VLA skills through an agent-style tool interface~\citep{rpent}. The brain repeatedly inspects intermediate results and maintains a cumulative multimodal conversation. In this system, the VLA is one tool in a richer program that also includes perception and geometry. The high-level model's invocation pattern follows task progress and tool results.

\paragraph{Robotic model serving.}
ROSA~\citep{rosa} coordinates GPU resources for multi-model robot pipelines involving planning, safety, and action models. Robion~\citep{robion} optimizes multi-robot VLA serving through GPU resource coordination, while VLA workload characterization~\citep{beyondllm} measures deadline-constrained inference and overlap between action and computation. Our study complements these systems by measuring the persistent context state and tool-return timing of a general-purpose agent brain alongside action-model deadline capacity.

\paragraph{Agent serving and KV-cache management.}
Continuum~\citep{continuum} optimizes KV preservation across tool gaps. CacheWise~\citep{cachewise} characterizes coding-agent prefixes and designs reuse-aware KV policies. EfficientAgent~\citep{efficientagent} models reuse working sets for concurrent agents. TAO~\citep{tao} incorporates tool readiness and cache value into scheduling; ScaleSim~\citep{scalesim} estimates invocation distance for multi-agent memory management. CONCUR~\citep{concur} shows KV-cache thrashing and congestion-aware agent admission. Our contribution measures how \emph{embodied} agent timing, action deadlines, and continuously replenished robot programs interact with these established mechanisms.

\paragraph{Working sets and recovery.}
Denning's working-set model~\citep{denning1968} explains cyclic cache replacement and admission-based relief. Research on metastable failures~\citep{bronson2021,huang2022} motivates testing whether a low-throughput state persists after a trigger has been withdrawn. We quantify delayed cache thrashing under sustained load and evaluate recovery following overload retraction.

\begin{table}[t]
\centering\footnotesize
\caption{Coverage of relevant serving research. Columns indicate robotic deadlines, agent key--value (KV) residency, sustained fleet replay, and analysis of both serving layers.}
\label{tab:related}
\setlength{\tabcolsep}{3pt}
\begin{tabular}{@{}lcccc@{}}
\toprule
 & Robot & Agent KV & Sustained & Brain+VLA \\
 & deadlines & residency & fleet replay & analysis \\
\midrule
ROSA~\citep{rosa} & $\checkmark$ & -- & $\checkmark$ & pipeline \\
CONCUR~\citep{concur} & -- & $\checkmark$ & batch jobs & -- \\
CacheWise~\citep{cachewise} & -- & $\checkmark$ & agent traces & -- \\
EfficientAgent~\citep{efficientagent} & -- & $\checkmark$ & agent traces & -- \\
This work & $\checkmark$ & $\checkmark$ & $\checkmark$ & $\checkmark$ \\
\bottomrule
\end{tabular}
\end{table}

\Cref{tab:related} positions the present evaluation among robotic model-serving and agent-cache systems. The central measured interaction is between a robot's control deadline and its brain's long-lived cache working set.

\section{Methodology}
\label{sec:method}

We combine traced embodied-agent programs, separate GPU profiling of the action and brain layers, and sustained replay of growing-context brain requests on a real serving engine. This design connects each model's measured serving cost to the dependency and timing structure of the original robot program.

\subsection{Task traces and timing conventions}
\paragraph{Collection.}
The brain is GPT-6-Astra (low reasoning effort), invoked through Codex in RPent~\citep{rpent}. The tools include rendering, geometric reasoning, scripted movement, and a LIBERO-tuned $\pi_{0.5}$ policy. A small paired no-SAM3 check produced 19/20 successes in each arm; the study uses the no-SAM3 setting. We execute ten tasks with two seeds in each of four LIBERO-PRO suites: 80 episodes, including 74 successes. Response-level logs record token use, brain elapsed time, per-tool environment steps, and VLA inference chunks. Responses are deduplicated by response ID.

\paragraph{What an inter-turn second measures.}
We distinguish three clocks. \emph{Brain request latency} is the recorded duration of one hosted Astra call, including the request's observed turnaround. \emph{Environment-step gap} is the number of steps advanced between consecutive brain calls divided by the 20\,Hz simulation control frequency. \emph{Simulator tool gap} is the corresponding original wall-clock gap, inflated by software rendering and inverse kinematics. These clocks describe complementary parts of a single robot program: the environment-step gap measures physical progression, and brain latency measures the time spent awaiting high-level decisions.

\begin{table}[t]
\centering\small
\caption{Timing of the 80-episode trace. Medians and 90th percentiles (P90) are in seconds. A paired response interval sums each logged Astra call with the preceding tool gap.}
\label{tab:timing}
\begin{tabular}{@{}lrr@{}}
\toprule
Trace timing quantity & Median & P90 \\
\midrule
Astra request latency & 7.78 & 13.84 \\
Environment-step gap & 1.00 & 3.10 \\
Simulator tool wall gap & 8.40 & 22.20 \\
Astra + environment gap & 9.40 & 15.38 \\
Astra + simulator wall gap & 17.96 & 32.77 \\
\bottomrule
\end{tabular}
\end{table}

\Cref{tab:timing} draws on all 1,862 brain request records and 1,782 within-episode gaps. For the paired intervals, we compute a sum per call before taking percentiles; medians of the separate components are not added. The hosted-brain request latency has a 48.26\,s maximum. Environment-step gaps are the timing inputs to the serving replay, where the local proxy's own measured response time replaces Astra request latency. The simulator wall-clock gap quantifies execution overhead; the environment-step gap determines the replay schedule.

\subsection{Hardware and two-layer profiling}
Our node has eight H800 80\,GB GPUs connected by NVLink, with GPUs 4--7 reserved for brain measurements. The brain proxy is Qwen3.6-27B in bfloat16 (BF16) at tensor parallelism four (TP=4), served by vLLM 0.17.1 with automatic prefix caching~\citep{vllm}. The NVIDIA driver is 535.161.08; \texttt{nvidia-smi} reports CUDA 12.3, while the Stage 6 user-space installation is PyTorch 2.10.0+cu128 (\texttt{torch.version.cuda}=12.8). A four-GPU all-reduce benchmark measured about 359\,GB/s bus bandwidth. The VLA uses five denoising steps per chunk. \Cref{tab:setup} collects the evaluated configuration and its measurement conventions.

\paragraph{Behavior and compute cost.}
Hosted Astra supplies the program's call sequence, context growth, token lengths, and tool gaps. Qwen3.6-27B supplies per-call compute cost on a controllable local serving engine. Re-tokenized length estimates had a median Astra/Qwen input-token ratio near 1.03. The replay uses random token IDs preserving input lengths and exact prefixes, isolating the cost and cache dynamics of the observed request pattern.

\begin{table}[t]
\centering\footnotesize
\caption{Primary serving configuration. BF16 denotes bfloat16, and TP denotes tensor parallelism. The CUDA version reported by \texttt{nvidia-smi} describes driver support; PyTorch specifies the user-space runtime.}
\label{tab:setup}
\begin{tabular}{@{}p{.22\columnwidth}p{.71\columnwidth}@{}}
\toprule
Hardware & 8$\times$ H800 80\,GB; brain study on GPUs 4--7 \\
Brain proxy & Qwen3.6-27B, BF16, TP=4, vLLM 0.17.1 \\
Runtime & NVIDIA 535.161.08 (driver CUDA 12.3); PyTorch 2.10.0+cu128 (user-space CUDA 12.8) \\
KV cache & Enabled; hybrid blocks of 784 tokens, with observed pools of 4,270 and 4,396 blocks \\
Trace brain & GPT-6-Astra, low effort, RPent/no-SAM3 \\
Action model & $\pi_{0.5}$; five denoising steps, five-step chunks \\
Robot control & 20\,Hz base; evaluated chunk budgets 0.25, 0.50, and 1.00\,s \\
Low-hit event & First 120\,s window with full-prefix hit rate below 80\% \\
\bottomrule
\end{tabular}
\end{table}

\subsection{Sustained and finite replay}
\label{sec:method:replay}
Each virtual robot issues dependent requests to the vLLM server using its previous turn's exact token prefix, the newly observed input length, and the trace's requested output length. After each response, the client advances by the recorded environment-step gap. In \emph{sustained} replay, each robot immediately draws a fresh episode upon completion. In the finite comparison, every robot runs two episodes and finishes.

We record full-prefix cache hits, completed request throughput, five-second Prometheus metrics, and client-side waiting. The full-prefix hit metric is conditioned on calls after the first turn of an episode. The onset measure is the first 120-second window with full-prefix hit rate below 80\%; runs without such a window contribute right-censored observations. Robot-level episode completion requires the final brain call in that episode's trace to finish; unfinished instances are recorded as censored.

\section{Workload Characterization}
\label{sec:workload}

Agentic execution produces a distinct two-layer workload: high-level turns follow perception, geometry, tool feedback, and physical task progress, while action chunks are issued only when selected by the brain. \Cref{tab:workload} summarizes the 80 collected episodes across four suites.

\begin{table*}[t]
\centering\small
\caption{The 80-episode RPent/Astra/$\pi_{0.5}$ workload. Statistics except success are per-episode medians. Input tokens are summed across an episode's brain turns; robot execution time is environment steps divided by 20\,Hz.}
\label{tab:workload}
\begin{tabular}{@{}lrrrrrr@{}}
\toprule
Suite & Success & Brain turns & Peak context & Total input & VLA chunks & Environment progression (s) \\
\midrule
\texttt{object\_swap} & 19/20 & 17.0 & 89.5k & 1.07M & 8 & 16.1 \\
\texttt{spatial\_swap} & 20/20 & 18.5 & 91.1k & 1.27M & 8 & 14.9 \\
\texttt{goal\_swap} & 19/20 & 19.0 & 95.5k & 1.32M & 15 & 21.8 \\
\texttt{10\_swap} (long) & 16/20 & 33.5 & 136.3k & 3.05M & 27 & 44.0 \\
\midrule
All & 74/80 & 19.0 & 95k & 1.31M & 12.5 & 19.5 \\
\bottomrule
\end{tabular}
\end{table*}

\paragraph{The brain participates frequently in task execution.}
The traces contain 1,862 brain turns and 216 VLA-tool entry points (186 \texttt{pi0\_pick} and 30 \texttt{pi0\_doubled}), within which the action model performs 1,582 inference chunks. Brain turns therefore outnumber VLA-tool entry points by 8.62$\times$ and individual VLA chunks by 1.18$\times$. These counts describe complementary levels of the tool hierarchy: an entry point may execute several chunks, and a brain turn may select perception, geometry, or multiple tools.

The episodes accumulate 2,309.05\,s of environment progression across 80 trajectories, corresponding to \textbf{0.806 brain turns per environment-execution second}. Across the object, spatial, goal, and long suites, this demand intensity is 1.010, 1.141, 0.820, and 0.640 turns per environment-execution second, respectively. The ratio measures \emph{brain-call demand relative to physical progression}; the actual request and response cycle also includes brain serving latency. In the hosted trace, the median Astra request latency is 7.78\,s and the median paired request-plus-environment interval is 9.40\,s (\cref{tab:timing}). These results establish the scale of reasoning time relative to robot motion.

\begin{figure}[t]
\centering\fig{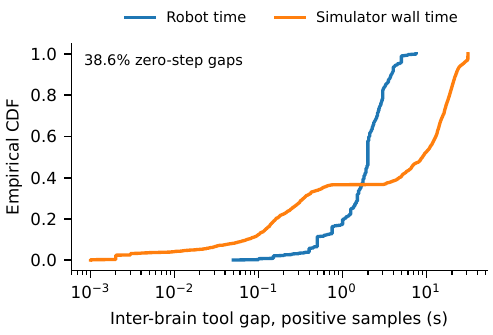}{\columnwidth}
\caption{Within-episode tool gaps between brain turns. The empirical cumulative distribution functions (ECDFs) show positive environment-step gaps and simulator wall-time gaps; 38.6\% of environment gaps advance zero steps. The plotted gap excludes the brain's own request latency.}
\label{fig:gaps}
\end{figure}

\begin{figure*}[t]
\centering\fig{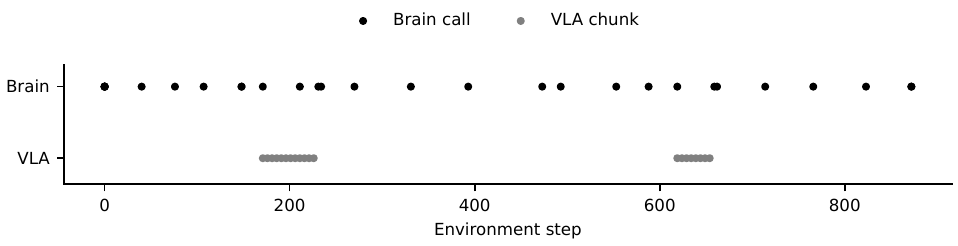}{0.88\textwidth}
\caption{Representative episode timeline using environment steps as a common horizontal axis. Brain turns and VLA inference chunks form separate, task-driven event streams.}
\label{fig:timeline}
\end{figure*}

\paragraph{Tool gaps create frequent opportunities for prefix reuse.}
Across 1,782 within-episode gaps, the environment-step gap has a 1.0\,s median, 3.1\,s 90th percentile, and 7.5\,s maximum. Exactly 688 gaps (38.6\%) advance zero environment steps, reflecting successive observations, reasoning turns, or non-motion tools. \Cref{fig:gaps} shows the empirical cumulative distribution function (ECDF) for positive gaps and reports the mass at zero separately. The simulator's corresponding median tool wall gap is 8.40\,s, reflecting rendering and inverse-kinematics overhead. \Cref{fig:timeline} illustrates the interleaved brain and action-chunk event streams along a representative episode's environment-step axis.

\begin{figure}[t]
\centering\fig{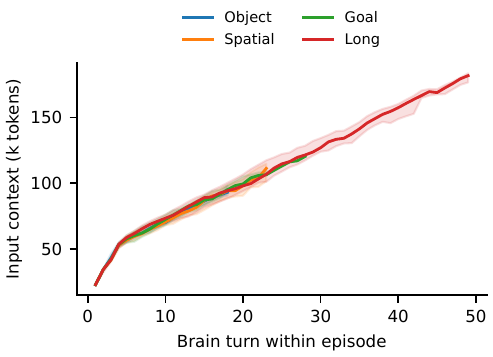}{\columnwidth}
\caption{Input context length over brain turns, with per-suite medians and interquartile ranges among episodes reaching each turn.}
\label{fig:context}
\end{figure}

\paragraph{Context grows throughout the episode.}
Every new prompt extends its predecessor; all 80 episodes retain cumulative context. Across 1,862 turns, median input length is 78.1k tokens, the 90th percentile is 131k, and the maximum is 214.7k. The long-task suite accumulates context more rapidly and reaches larger prompts (\cref{fig:context}). An individual call contains a median of 21 image references and up to 141. This growing context persists across successive tool returns.

\begin{figure*}[t]
\centering\fig{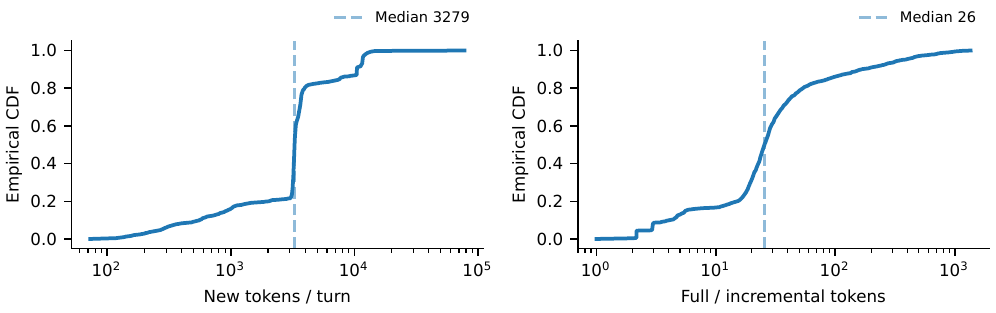}{0.87\textwidth}
\caption{Per-turn incremental input length and the full-prompt/incremental-token ratio. These distributions quantify how much prefill work retained key--value cache state can save.}
\label{fig:newtok}
\end{figure*}

\paragraph{A small increment can reactivate a large prefix.}
The median turn appends 3,279 input tokens (90th percentile 10,475) and generates 144 output tokens. The median ratio of full prompt length to newly appended input is 25.5$\times$. It quantifies the potential redundant prefill work following complete prefix eviction. \Cref{fig:newtok} shows the distribution of incremental lengths and recomputation ratios; the realized vLLM costs are measured in \cref{sec:brain}.

\paragraph{Serving implication.}
Robot count determines how many long-lived contexts compete for memory, while the tool-return process determines how often those contexts must be revisited. The resulting serving demand is governed by \emph{simultaneous, frequently reused KV working sets} and by the inference time required to advance each robot program.

\section{VLA Capacity Depends on the Control Deadline}
\label{sec:vla}

\begin{figure}[t]
\centering\fig{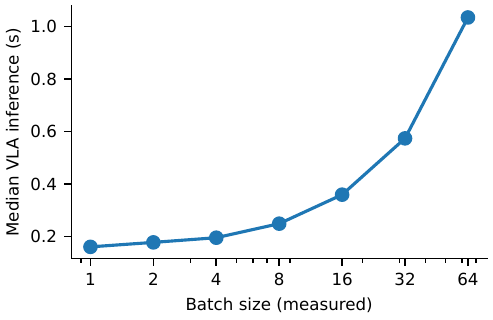}{\columnwidth}
\caption{Measured $\pi_{0.5}$ inference latency by batch size. The seven plotted sizes are directly measured medians over 50 runs each, preceded by 10 warm-up runs per shape.}
\label{fig:vla}
\end{figure}

The VLA serves short, latency-sensitive action chunks. In the measured $\pi_{0.5}$ configuration, batch-one inference takes 160\,ms at the median; vision encoding and vision--language model (VLM) prefill each take about 19\,ms, followed by five denoising steps of roughly 23\,ms each. Nsight profiling shows substantial idle time during small-batch denoising, consistent with launch and synchronization overhead. Batching raises arithmetic utilization and latency: the observed median is 249\,ms at batch eight and 1.03\,s at batch 64. \Cref{fig:vla} reports all seven measured batch sizes.

\paragraph{Deadline-qualified robot capacity.}
We drive a first-come, first-served (FCFS) batching server with virtual robots executing five-step chunks at a fixed control frequency. The policy can prefetch action chunks and maintains at most one outstanding request per robot. A control tick is \emph{starved} when its action buffer is empty. We define capacity at chunk deadline $D=5/f$ as the largest tested robot count $N$ for which both repeated runs have at most 1\% starved ticks. All values in \cref{tab:vlacap} use continuous demand, five-step prefetch, and zero network round-trip time (RTT).

\begin{table}[t]
\centering\small
\caption{Deadline-qualified VLA capacity under a 1\% starvation service-level objective (SLO), with five-step prefetch and zero round-trip time (RTT). Capacity is the largest tested robot count passing the $\leq1\%$ starvation target in both repetitions. One-GPU tests cover $N\in\{1,2,4,8,16,32\}$; two-GPU tests cover $N\in\{8,16,32\}$.}
\label{tab:vlacap}
\begin{tabular}{@{}lcc@{}}
\toprule
Chunk deadline & 1 GPU & 2 GPUs \\
\midrule
0.25\,s (20\,Hz) & 1 & no passing test at $N\geq8$ \\
0.50\,s (10\,Hz) & 8 & 16 \\
1.00\,s (5\,Hz) & $\geq32$ & $\geq32$ \\
\bottomrule
\end{tabular}
\end{table}

\begin{figure*}[t]
\centering\fig{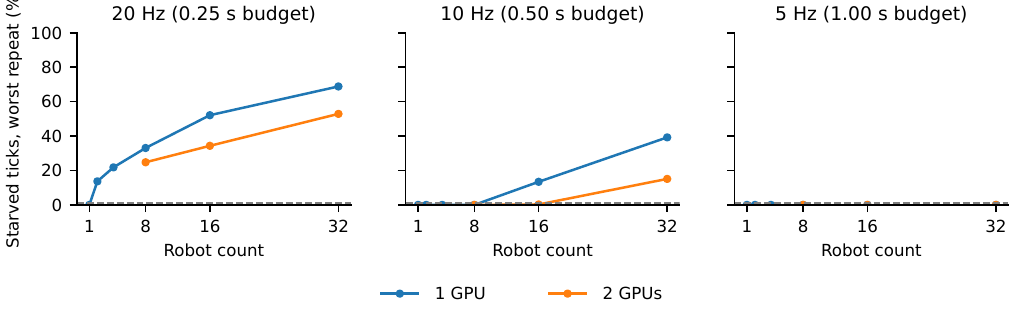}{0.94\textwidth}
\caption{Control-tick starvation as robot count grows. Each point represents the worse of two runs; dashed horizontal lines mark the 1\% service-level objective. Each panel uses its labeled chunk deadline and replica configuration.}
\label{fig:vlacap}
\end{figure*}

\Cref{fig:vlacap} shows the measured starvation fraction versus robot count for each deadline and replica configuration.

The feasible operating region changes sharply with deadline. One H800 accommodates one robot with a 0.25\,s budget, eight at 0.50\,s, and all 32 tested robots at 1.00\,s. Two GPUs support 16 robots at 0.50\,s. In the 0.25\,s two-GPU configuration, the smallest measured fleet contains eight robots and already violates the target; the lower-count capacity remains unmeasured.

The test also reveals the scheduling mechanics: sending the next request after exhausting the preceding chunk creates unavoidable waiting, while poorly aligned arrivals limit batch formation. Deadline-qualified capacity therefore depends on arrival phase and action buffering alongside peak GPU arithmetic throughput.

\paragraph{Exposing the brain-side capacity constraint.}
Tight action deadlines make VLA latency the immediate scale limit. A longer chunk budget or additional effective VLA replicas increases the action layer's admissible robot count, exposing the second limit: the number of active brain contexts that can retain their KV working sets. Both constraints determine the feasible deployment envelope.

\section{The Brain's KV Working Set}
\label{sec:brain}

\begin{figure}[t]
\centering\fig{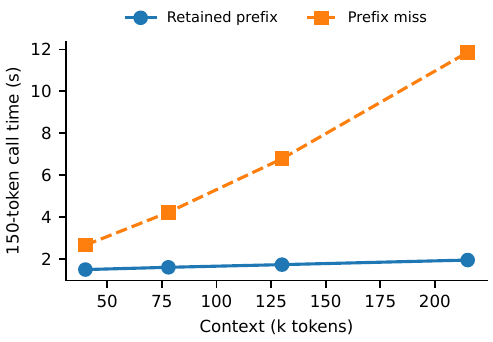}{\columnwidth}
\caption{Estimated 150-token brain call duration from measured TTFT and batch-one decoding costs, under retained and evicted KV prefixes.}
\label{fig:callcost}
\end{figure}

\paragraph{Cache retention controls request cost.}
We profile Qwen3.6-27B in bfloat16 at tensor parallelism four on H800 GPUs. Its hybrid attention architecture uses 784-token vLLM prefix-cache blocks. For an evicted prefix, the median time to first token (TTFT) grows from 1.34\,s with a 40k-token context to 10.26\,s with 215k tokens. Retaining the preceding prefix reduces median TTFT to 0.18--0.35\,s when a turn appends 3,300 new tokens. Batch-one decoding costs roughly 8.9--10.7\,ms/token. \Cref{tab:braincost} reports the measured TTFT and its combination with 149 decode intervals; \cref{fig:callcost} visualizes the resulting complete-call cost.

\begin{table}[t]
\centering\small
\caption{Brain proxy cost at tensor parallelism four. Time to first token (TTFT) is measured; a 150-output-token call combines TTFT with 149 measured batch-one decode intervals. The retained-prefix case adds 3.3k tokens. All entries are seconds.}
\label{tab:braincost}
\begin{tabular}{@{}lrrrr@{}}
\toprule
Context & 40k & 78k & 130k & 215k \\
\midrule
TTFT, evicted & 1.34 & 2.85 & 5.32 & 10.26 \\
TTFT, retained & 0.18 & 0.22 & 0.27 & 0.35 \\
Call, evicted & 2.66 & 4.23 & 6.78 & 11.86 \\
Call, retained & 1.50 & 1.61 & 1.73 & 1.95 \\
\bottomrule
\end{tabular}
\end{table}

At 130k tokens, retaining the prefix reduces a 150-token call from about 6.8\,s to 1.7\,s. The survival of a preceding conversational KV state is therefore a first-order determinant of both compute expenditure and the waiting time of the robot program.

\begin{figure*}[t]
\centering\fig{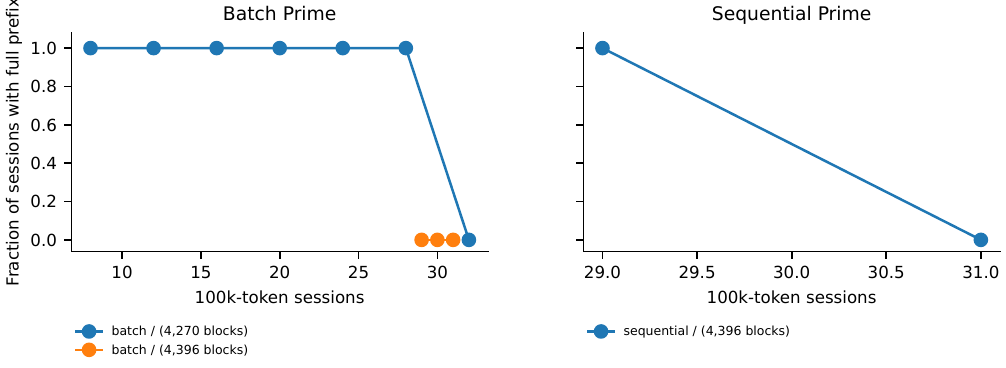}{0.94\textwidth}
\caption{Full-prefix survival under cyclic, ordered revisits. Batch priming and sequential priming are shown separately, and each curve identifies the underlying KV-cache block pool.}
\label{fig:cliff}
\end{figure*}

\paragraph{A sharp retention boundary that depends on priming.}
We prime $N$ independent 100k-token sessions and revisit them in the same cyclic order, appending 3.3k tokens per revisit. \Cref{fig:cliff} distinguishes three measurement configurations:
\begin{itemize}
\item With 4,270 cache blocks and batch priming, $N=28$ retains all prefixes and $N=32$ loses all full-prefix reuse.
\item With 4,396 blocks and batch priming, the tested $N=29$--31 cases lose every full prefix.
\item With 4,396 blocks and sequential priming, $N=29$ retains every prefix and $N=31$ loses every full prefix.
\end{itemize}

The abrupt loss of reuse follows the classical \emph{least-recently-used} (LRU) working-set mechanism: refilling one evicted session can displace the prefix needed by the next~\citep{denning1968}. Priming order and effective cache capacity shift the observed threshold. The sustained-fleet experiments next expose the dynamic effect of this replacement mechanism under heterogeneous contexts and tool-return times.

\section{Persistent Thrashing Under Fleet Load}
\label{sec:fleet}

We now run the collected embodied-agent programs as a sustained fleet. Each of $N$ virtual robots issues dependent requests to a single tensor-parallel vLLM brain server and immediately samples a new episode upon completing its previous one. Heterogeneous context sizes, short tool gaps, and continuous admission of new episodes produce a persistent, fluctuating KV working set.

\begin{figure*}[t]
\centering\fig{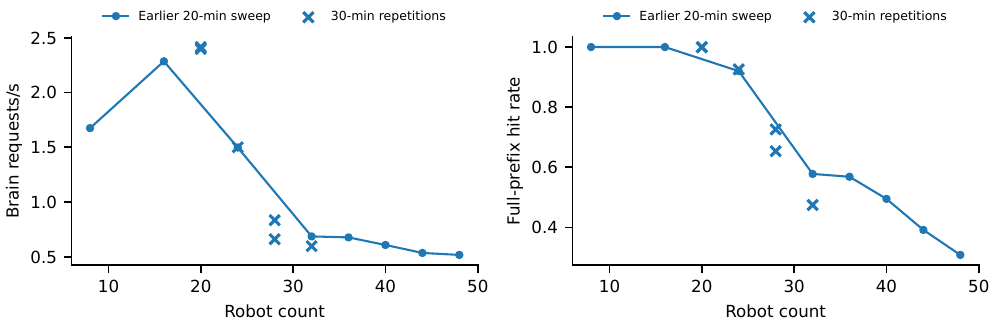}{0.94\textwidth}
\caption{Sustained-fleet request throughput and full-prefix hit rate. Circles represent the historical 20-minute sweep across 8--48 virtual robots; crosses represent 30-minute replication runs. The two series retain their original durations and seed provenance.}
\label{fig:fleet}
\end{figure*}

\paragraph{More robots produce less completed work beyond the retention boundary.}
\Cref{fig:fleet} combines the 8--48 robot capacity sweep with 30-minute replication runs at $N=20$--32. At 20 robots, two 30-minute runs achieve 2.40--2.42 completed requests/s with approximately 99.95\% non-initial full-prefix hits. At 28 robots, two runs deliver 0.66--0.83 requests/s; at 32, the measured request throughput is about 0.60/s with 47.5\% aggregate full-prefix hits. An earlier unrestricted 48-robot run achieves 0.52 requests/s. Relative to 20 robots, 32 robots yield roughly fourfold fewer completed brain requests per second.

\begin{figure}[t]
\centering\fig{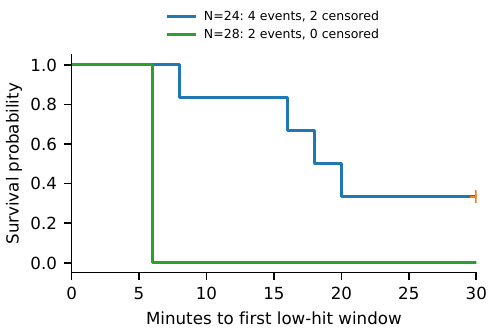}{\columnwidth}
\caption{Kaplan--Meier survival estimate for the first 120-second window with full-prefix hit rate below 80\%. The $N=24$ group has four events and two right-censored runs; $N=28$ has two observed events.}
\label{fig:survival}
\end{figure}

\paragraph{The onset of cache thrashing is delayed and load-dependent.}
The 32-robot server initially benefits from retained prefixes and then enters a prolonged low-hit operating regime. In six 30-minute $N=24$ trials, four first cross the 80\% hit-rate threshold at 480, 960, 1080, and 1200\,s; the other two runs are right-censored at 30 minutes. Both available $N=28$ trials cross at 360\,s. \Cref{fig:survival} characterizes \emph{time to the first 120-second low-hit window}, including its observed variability and censoring.

\begin{figure*}[t]
\centering\fig{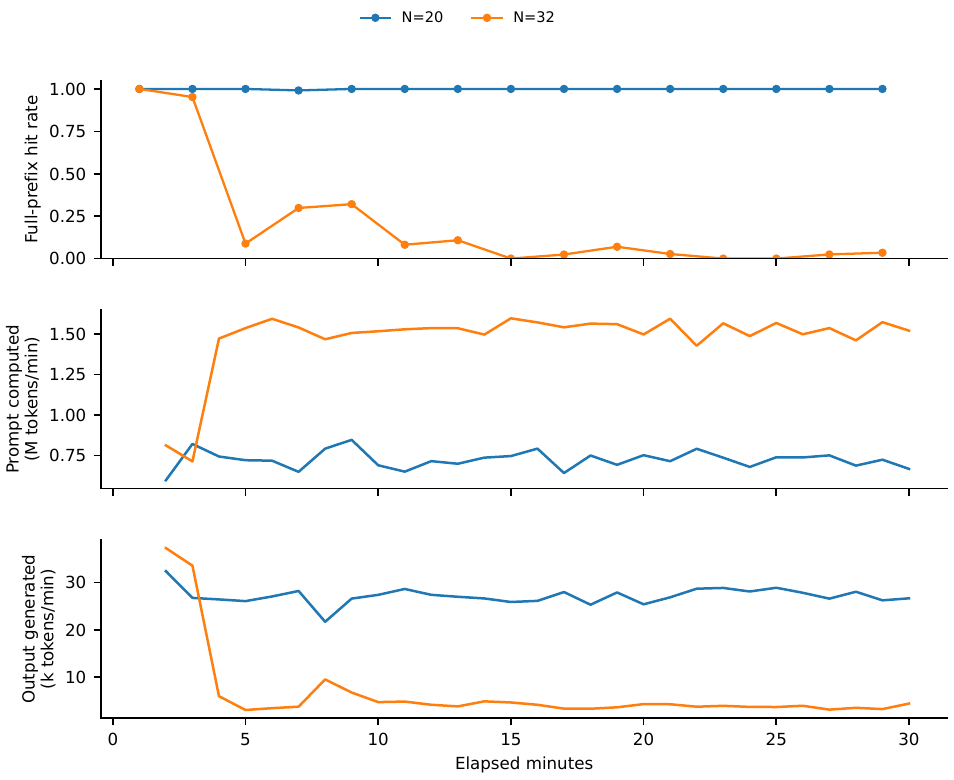}{0.90\textwidth}
\caption{Full-prefix hit rate (120-second windows) and vLLM token computation (60-second windows) for healthy $N=20$ and thrashing $N=32$ runs. Prompt computation rises while useful generated-token throughput declines.}
\label{fig:timeseries}
\end{figure*}

\paragraph{Repeated prefill consumes the useful generation budget.}
The measured vLLM performance counters reveal the mechanism. A stable $N=20$ run locally computes approximately 0.7 million prompt tokens per minute and generates about 27k output tokens/min. In the late $N=32$ thrashing regime, locally computed prompt tokens rise to roughly 1.5 million/min while generated output falls to about 4k tokens/min. \Cref{fig:timeseries} reports the directly reconstructed 60-second counter differences. Greater prefill demand coincides with far less output available to advance robot programs.

This behavior is consistent with a KV-residency--prefill feedback loop. Waiting extends the interval before a session can reuse its prefix; recency-based eviction then increases its chance of a costly refill. That refill delays other sessions, exposing yet more contexts to eviction. The runtime's \texttt{kv\_cache\_usage\_perc} measures occupancy of running-request storage, while retained, reclaimable KV also governs reuse. Full-prefix survival and computed prompt tokens therefore provide complementary signals for capacity control. The mechanism connects to CONCUR's analysis of concurrent LLM agents~\citep{concur}.

\paragraph{Recovery and arrival history.}
Three complementary tests probe the dynamics. In the two overload/retraction runs (E2a), hit rate recovered after the overload was removed, including a run that recovered within about six minutes. Three operational $N=22$ arrival-path replays (E2b) achieved 2.36, 2.38, and 1.73 requests/s for cold-start, ramp-up, and ramp-down constructions. The different histories produce distinct serving trajectories. In a drain/release test at $N=28$ (E2c), the active set first reaches 16 near 941\,s, after the initial 600-second reduction request; hit rate then improves. The analysis uses the effective draining point reconstructed from active-episode intervals.

\paragraph{Continuously replenished fleets sustain the working set.}
CONCUR studies finite agent batches, whose resident contexts naturally decline as jobs complete~\citep{concur}. Our finite experiment (E5) runs two episodes per robot: all 48 episodes complete at $N=24$, and all 56 complete at $N=28$. At $N=28$, the finite replay eventually returns to full-prefix hits as active work drains. Under the same seed, continuous replenishment maintains workload pressure and the late low-hit regime (Appendix~\ref{sec:appendix}). The finite-to-sustained comparison identifies the importance of \emph{long-term active working-set occupancy} in robotic fleet serving.

\section{Working-Set Admission Restores Progress}
\label{sec:mitigation}

Working-set admission limits simultaneously active processes so that reusable memory state survives until each process returns~\citep{denning1968}. CONCUR applies congestion-aware concurrency control to generic LLM agents~\citep{concur}. We evaluate episode-boundary admission for the embodied workload: a cap $K$ permits at most $K$ active episodes, while additional robots wait to begin another episode. This directly controls the population of long-lived brain contexts.

\begin{figure*}[t]
\centering\fig{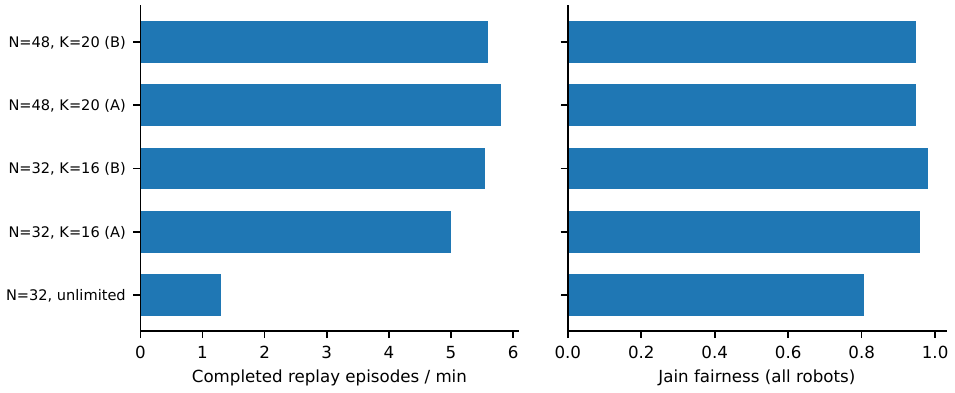}{0.92\textwidth}
\caption{Completed fixed-trace robot programs per minute and Jain's fairness index across all target robots. Each horizontal bar is one independent run; A and B distinguish separate runs at the same setting. Fully replayed episodes count toward throughput, while unfinished instances are censored.}
\label{fig:robotmetrics}
\end{figure*}

\paragraph{Complete robot-program throughput improves as well.}
We credit an episode only after its final original brain call completes and count all target robots, including those with zero completions, in Jain's fairness index. At $N=32$, unrestricted replay completes 39 episodes (1.30/min; fairness 0.806). Two $K=16$ runs complete 100 and 111 (5.00 and 5.55/min; fairness 0.959 and 0.980). At $N=48$, two $K=20$ runs complete 116 and 112 (5.80 and 5.60/min; fairness 0.947 in both). \Cref{fig:robotmetrics} shows that the server-side throughput recovery translates into more completed agent programs across the fleet.

\paragraph{Request throughput returns to the cache-resident regime.}
At $N=32$, unrestricted replay achieves 0.598 completed requests/s. With $K=16$, two trials achieve 2.235 and 2.253 requests/s, each with 100\% non-initial full-prefix hits. The time series shows a stable high-hit regime under admission (Appendix~\ref{sec:appendix}). At $N=48$, the evaluated caps $K=12,16,20,24$ yield 2.027, 2.269, 2.377 (two-run mean), and 2.132 requests/s, respectively. The highest observed value occurs at $K=20$ (\cref{tab:admissionscan}; Appendix~\ref{sec:appendix}), improving on the historical unrestricted $N=48$ throughput of about 0.52 requests/s by approximately 4.6$\times$.

\begin{table}[t]
\centering\small
\caption{Static active-episode admission for 48 virtual robots. The $K=20$ results average two valid runs; other caps each have one valid run. Full-hit percentages exclude the first brain call of each episode.}
\label{tab:admissionscan}
\begin{tabular}{@{}rrrr@{}}
\toprule
Cap $K$ & Requests/s & Full-hit (\%) & Runs \\
\midrule
12 & 2.027 & 100.00 & 1 \\
16 & 2.269 & 100.00 & 1 \\
20 & 2.377 & 99.95 & 2 \\
24 & 2.132 & 98.01 & 1 \\
\bottomrule
\end{tabular}
\end{table}

\paragraph{Congestion signals govern adaptive admission.}
We also tested an independently implemented, CONCUR-inspired additive-increase/multiplicative-decrease (AIMD) heuristic. It reads KV usage, prefix hits, and queue depth every 30\,s and updates the active-episode cap at episode boundaries. At $N=32$ and $N=48$, it achieved 1.280 and 1.161 requests/s with 88.4\% and 85.8\% full-prefix hits. All 80 control records show \emph{zero multiplicative-decrease decisions}: the rule requires high \texttt{kv\_cache\_usage\_perc} in combination with lower cache hits, leaving the cap as high as 27 during thrashing. The resulting control trajectory demonstrates that admission telemetry should directly reflect \emph{reusable prefix survival}; detailed controller trajectories appear in Appendix~\ref{sec:appendix}.

\paragraph{Operational implication.}
Fleet population and concurrently admitted working-set size are separate control variables. The first drives the backlog of programs seeking service; the second determines whether the engine can retain those programs' KV state. Episode completion and per-robot fairness determine how efficiently a serving policy advances the overall robot workload.

\begingroup
% Keep the disclosure with the closing text without shrinking the body font.
\makeatletter
\renewcommand\section{\@startsection{section}{1}{\z@}%
  {-1.5ex \@plus -.5ex \@minus -.2ex}{1ex \@plus .2ex}{\normalfont\Large\bfseries}}
\renewcommand\paragraph{\@startsection{paragraph}{4}{\z@}%
  {1ex \@plus .2ex \@minus .2ex}{-1em}{\normalfont\normalsize\bfseries}}
\makeatother
\section{Discussion and Design Implications}
\label{sec:discussion}

\paragraph{A joint serving-capacity envelope.}
The action model and brain impose different resource constraints. Action serving is deadline-qualified: a robot is served only if its required chunks arrive before the action buffer runs dry. Brain serving is working-set-qualified: reused conversational prefixes must remain resident across tool waits. With the measured 0.25\,s VLA deadline, an action-model GPU satisfies the starvation objective for one robot; at a 1.0\,s deadline it serves at least 32 in the measured range. In contrast, the studied brain proxy's high-hit regime is near 20 concurrently active contexts. Practical sizing follows a joint envelope that varies with deadline, replica count, brain tensor parallelism, and admitted sessions. The fleet stability study concentrates repeated 30-minute trials around $N=20$--32; single-run admission settings remain visibly identified in the cap sweep.

\paragraph{Useful progress is the fleet-level objective.}
Large VLA batches can increase arithmetic throughput while delaying action chunks. High brain concurrency can intensify compute utilization while reducing completed agent requests through repeated prefill. The relevant performance target is therefore \emph{deadline-qualified action progress and completed robot programs}, supported by server metrics such as prefix hits, computed prompt tokens, generated output tokens, and queue depth. Our episode-completion measurements show that a smaller resident working set can accelerate fleet-wide progress.

\paragraph{Opportunities for robot-aware scheduling.}
Robot tools expose action-chunk lengths, remaining environment steps, and progress markers. These signals can estimate the timing of the next brain revisit and enable cache-residency-aware scheduling and eviction. Such policies extend existing ideas on tool-aware KV retention and invocation distance~\citep{continuum,tao,scalesim}. Our measurements identify the robot execution signals and the sustained-load operating region where these techniques can be assessed.

\paragraph{Scope and next measurements.}
The workload comes from one RPent configuration, four LIBERO-PRO suites, and Astra at low reasoning effort. The serving replay follows each program's token lengths, exact shared-prefix relationships, and environment-step gaps using Qwen3.6-27B as the cost proxy. A fully integrated brain--VLA deployment with actual multimodal message processing and shared-cluster interference is the next validation step. The completed replay-episode metric measures service progress through a recorded program; closed-loop task success under altered response latencies remains a future evaluation. Finally, the overload/retraction experiments recovered after the trigger was withdrawn, establishing delayed, sustained-load thrashing as the measured failure mode. This separation of onset, duration, and recovery supports precise operating-point design.

\section{Conclusion}
\label{sec:conclusion}

Tool-driven agentic robots create a serving workload defined by frequent, long-context brain calls and deadline-sensitive action chunks. Our measurements connect this workload to two distinct capacity constraints and show how the limiting resource moves with control budgets and admitted KV working sets. Under sustained fleet replay, cascading prefix misses turn useful generation into repetitive prefill and reduce completed work. Episode-level admission restores cache residency, brain throughput, and complete trace-program throughput. These results offer an empirical foundation for capacity planning and future brain--VLA co-scheduling in shared robot-serving clusters.

\paragraph{Acknowledgments.} The author thanks Lingxi Xie, Yunning Cao, Pengfei Chen, Kaiwen Duan, Jiannan Ge, Xinyue Huo, Jiacheng Shao, and Qi Tian.
Generative AI tools were used to assist with software development, data analysis, and manuscript preparation. The author takes responsibility for the reported results and final manuscript.

\par\endgroup

\FloatBarrier
\clearpage
\onecolumn
\begin{multicols}{2}
\small\setlength{\bibsep}{1pt}
\bibliographystyle{unsrtnat}
\bibliography{references}
\end{multicols}
\appendix
\section{Trace Timing Audit}\label{sec:timingaudit}
\Cref{tab:timingsuite} reports the suite-level measurements underlying the time decomposition in \cref{tab:timing}. Here, the demand intensity is the ratio of brain turns to cumulative environment progression; each suite's environment time is the sum of its recorded steps divided by 20\,Hz.

\begin{table}[ht]
\centering\small
\caption{Trace-time normalization by suite. The paired median is calculated over within-episode intervals by adding each recorded Astra request latency to its preceding environment-step gap.}
\label{tab:timingsuite}
\begin{tabular}{@{}lrrrrr@{}}
\toprule
Suite & Turns & Env. time (s) & Turns/s$_{\rm env}$ & Astra (s) & Paired (s) \\
\midrule
Object & 336 & 332.80 & 1.010 & 8.54 & 10.21 \\
Spatial & 385 & 337.40 & 1.141 & 7.33 & 8.59 \\
Goal & 417 & 508.35 & 0.820 & 8.55 & 10.44 \\
Long & 724 & 1130.50 & 0.640 & 7.22 & 9.05 \\
\midrule
All & 1862 & 2309.05 & 0.806 & 7.78 & 9.40 \\
\bottomrule
\end{tabular}
\end{table}

The original hosted-Astra log contains 1,862 request durations and 1,782 within-episode tool gaps. The summary statistics were reconstructed from per-turn execution records, with source checksums and derivation steps retained in the project provenance materials. The hosted request duration includes its observed turnaround, whereas the environment-step clock records only physical progress. In the brain-proxy replay, serving time comes from measured Qwen inference and scheduling on vLLM; the environment-step gap determines the subsequent tool wait.

\clearpage
\section{Additional Fleet Diagnostics}\label{sec:appendix}
\begin{figure}[ht]
\centering
\begin{minipage}[t]{0.47\textwidth}
\centering\fig{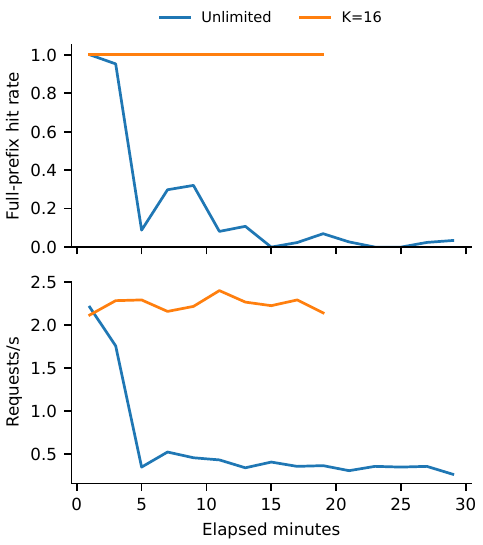}{0.77\linewidth}
\caption{At $N=32$, an episode cap $K=16$ maintains high-prefix-hit serving and request throughput; the two curves use independent seeds.}
\label{fig:admission}
\end{minipage}\hfill
\begin{minipage}[t]{0.47\textwidth}
\centering\fig{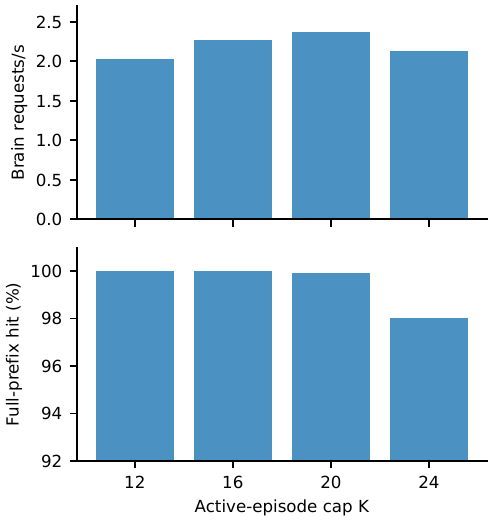}{0.77\linewidth}
\caption{Static admission sweep at $N=48$, showing completed brain requests/s and full-prefix hit rate. $K=20$ averages two valid trials.}
\label{fig:scan}
\end{minipage}
\end{figure}

\begin{figure}[ht]
\centering
\begin{minipage}[t]{0.47\textwidth}
\centering\fig{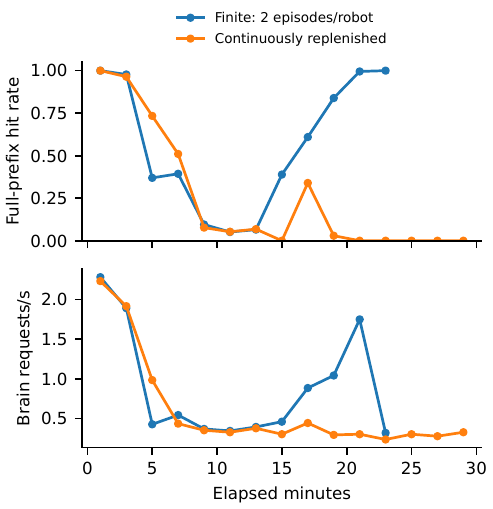}{0.77\linewidth}
\caption{Same-seed $N=28$ finite and sustained fleet replay: cache hits and completed requests by two-minute window. Finite programs end after two episodes per robot.}
\label{fig:finite}
\end{minipage}\hfill
\begin{minipage}[t]{0.47\textwidth}
\centering\fig{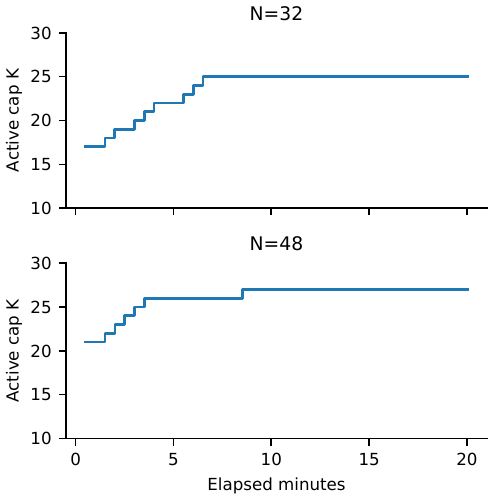}{0.77\linewidth}
\caption{Active-episode cap of the independently implemented, CONCUR-inspired additive-increase/multiplicative-decrease (AIMD) heuristic. All 80 control decisions show zero decreases.}
\label{fig:adaptive}
\end{minipage}
\end{figure}

The following time-series results provide supplementary evidence for the finite-versus-sustained comparison and the adaptive controller's behavior. \Cref{fig:admission,fig:scan,fig:finite,fig:adaptive} present the admission, finite-fleet, and adaptive-control diagnostics.

\par\smallskip
\noindent The time-series panels below show the working-set dynamics as finite robots finish their programs and as an adaptive controller increases its cap.

\paragraph{Reconstruction details.}
The E2c drain reached 16 active episodes at an estimated $940.565$\,s based on the recorded active-episode intervals. Later control instrumentation captures explicit request, effect, and release timestamps. Finite-batch completion counts require all turns in a recorded episode to finish. The E3 survival event is the first 120-second full-prefix-hit window below 80\%, with right-censoring at the end of observation.

\end{document}